\documentclass[a4paper, 10pt, conference]{ieeeconf}      
\usepackage{FG2025}

\FGfinalcopy 
\usepackage{comment}

\IEEEoverridecommandlockouts                              
\usepackage{graphics} 
\usepackage{epsfig} 
\usepackage{mathptmx} 
\usepackage{times} 
\usepackage{amsmath} 
\usepackage{amssymb}  
\usepackage{lipsum}
\usepackage{booktabs}
 \usepackage{multirow} 
\def\FGPaperID{154} 

\title{\LARGE \bf
Salience Adjustment for Context-Based Emotion Recognition
}

\author{\parbox{16cm}{\centering
    {\large Bin Han and Jonathan Gratch}\\
    {\normalsize
    Computer Science, University of Southern California, Los Angeles, USA\\}}
}

\begin{document}

\ifFGfinal
\thispagestyle{empty}
\pagestyle{empty}
\else
\author{Anonymous FG2025 submission\\ Paper ID \FGPaperID \\}
\pagestyle{plain}
\fi
\maketitle

\begin{abstract}
Emotion recognition in dynamic social contexts requires an understanding of the complex interaction between facial expressions and situational cues.
This paper presents a salience-adjusted framework for context-aware emotion recognition with Bayesian Cue Integration (BCI) and Visual-Language Models (VLMs) to dynamically weight facial and contextual information based on the expressivity of facial cues.
We evaluate this approach using human annotations and automatic emotion recognition systems in prisoner’s dilemma scenarios, which are designed to evoke emotional reactions.
Our findings demonstrate that incorporating salience adjustment enhances emotion recognition performance, offering promising directions for future research to extend this framework to broader social contexts and multimodal applications.
\end{abstract}

\section{INTRODUCTION}

Automatic expression recognition traditionally treats facial expressions as signifying the emotional state of the expresser. 
Recently, there has been growing appreciation that observer perceptions differ from self-reported emotions~\cite{barrett2011context} (people can seem happy when experiencing negative emotions~\cite{ansfield2007smiling,hladky2024modeling}), yet these (mis)perceptions are crucial for explaining human social behavior. 
Perceived emotions build trust and cooperation~\cite{boone2003emotional,van2004interpersonal}, shape partner decisions~\cite{de2014reading}, and help others to regulate their own emotions~\cite{niedenthal2002adult}, regardless of the expresser's true feelings. Automatically recognizing perceived emotion can improve theories of human social behavior~\cite{heesen2024impact} and human-machine interaction~\cite{de2014humans}.

Research into how observers interpret facial expressions highlights the crucial role of context. Interpretations from face alone differ dramatically from how expressions are interpreted in the context of other modalities~\cite{poria2018meld}, background faces~\cite{masuda2008placing}, co-occurring actions~\cite{aviezer2008angry}, or recent events~\cite{han2024knowledge}. Automated approaches to such “context-based emotion recognition” include standard machine learning~\cite{kosti2019context,lee2019context} and methods grounded in psychological theory~\cite{Klinger2023appraisal,mortillaro2012advocating,han2024knowledge}. 
In this paper, we focus on psychologically motivated approaches as our interest is in uncovering fundamental mechanisms that shape emotion perception, though we argue that such approaches can still yield state-of-the-art performance.

Bayesian Cue Integration (BCI) is a prominent psychological theory of how people infer emotion in context~\cite{ong2015affective,goel2024face}. 
From the perspective of automatic recognition, BCI is a cognitively-plausible late-fusion approach that can be applied to any state-of-the-art context-free emotion recognition method. It works by post-processing its output into predictions that better align with context-based annotations (see~\cite{han2024knowledge}). BCI argues people form separate emotion judgments from expressions and from situations, then integrate them with Bayesian inference. The theory behind BCI has been validated across a diverse range of social situations~\cite{saxe2017formalizing,tenenbaum2011grow, ong2015affective,zaki2013cue}. 
While the theory highlights the importance of face and situation in determining perception, it also suggests that additional factors should be incorporated into the model. 
For instance, a recent study found that contextual information may dominate facial information in certain cases~\cite{goel2024face}, suggesting the need to weigh the relative contribution of face versus situational context.

In this paper, we propose that the visual salience of a facial expression is an important factor that determines the relative importance of face versus context when observers form context-based emotion judgments. We first discuss prior psychological research that suggests that strong facial expressions capture attention. We next examine how facial and contextual cues interact to shape human emotion perception in a social task with real-world consequences. 

We make several contributions to Bayesian Cue Integration and its application to automatic expression recognition. First,  we find evidence that observers place greater weight on facial expressions than contextual cues when facial movement is salient. Second, we show that BCI's accuracy improves by incorporating expression salience. Third, we show that adjusting for expression salience enhances the accuracy of automatic recognition systems inspired by BCI. Finally, we demonstrate that even non-BCI approaches, such as the use of Vision-Language Models, achieve higher accuracy by adjusting for visual salience.

\section{Expression Salience}
\label{sec:salience}

Several studies find that emotional faces capture visual attention. Emotional faces ``pop out" of a scene~\cite{calvo2008visual}.
Strong expressions, especially smiles~\cite{calvo2008detection} and frowns~\cite{fox2000facial}, 
attract attention and are recognized faster than more neutral faces or other visual information. The urge to attend to strong expressions is difficult to suppress, meaning that expressive faces cannot be ignored as effectively as other sources of information~\cite{blagrove2014ignoring}. Previous studies on BCI also found that the relative contribution of face and context varies, but did not incorporate a mechanism to address this variance~\cite{goel2024face}.

Building on these findings, we hypothesize that facial expressivity is one mechanism that modulates the relative importance of facial versus situational cues when forming an overall emotion judgment:

\begin{itemize} \item \textit{Expression-Salience Hypothesis}: When facial expressiveness is low, perceivers give more weight to situational cues; when facial expressiveness is high, perceivers prioritize facial cues. \end{itemize}

To test this hypothesis, we analyzed how the salience of the expression influences the relative attention to facial versus situational cues in a social task with financial consequences. We first verify the hypothesis using purely human judgments before exploring whether the hypothesis affects the accuracy of automatic recognition. 

\subsection{Dataset: USC Split-Steal}
We examine the impact of expression salience using the USC Split-Steal corpus~\cite{lei2023emotional} as this has been previously used to illustrate the relevance of BCI to automatic expression recognition~\cite{han2024knowledge}.
The Split-Steal corpus contains videos and meta-data on participants engaged in a 10-round prisoner’s dilemma~\cite{rapoport1965prisoner}. This task creates tension between cooperation and competition. On each round, participants can choose to cooperate (C) by offering to split a pot of lottery tickets, or they can defect (D) by trying to steal the entire pot. Their earnings are determined by the joint decision (i.e., they can steal the entire pot if they choose D and their partner chooses C).  The game creates an incentive to cooperate but a temptation to steal and fear of being exploited. Prior research shows that people pay attention to their partner's emotional expressions after each round to predict their partner's intentions and determine their own actions~\cite{de2021emotion}.  

We analyzed 100 videos from the Split-Steal corpus that had previously been annotated with context-free and context-based labels~\cite{han2024knowledge}  -- 25 from each joint outcome (CC, CD, DC, DD). These illustrate a player's facial reaction upon learning the outcome in a round (7-second video). Human annotators (N=141) provided 20 ratings per video of valence (5-point Likert scale~\cite{bradley1994measuring}), and Basic Emotion (anger, disgust, fear, joy, sadness, surprise or neutral). Basic emotion labels were chosen as these have been used in prior psychological research on the prisoner's dilemma (e.g.,~\cite{de2021emotion}) 
and BCI~\cite{ong2015affective}, but we also add valence as a second measure of emotion.

Annotations were collected under three conditions:


\begin{itemize}
    \item \textbf{Context-free (Only Face)}: 
Annotators only saw the video reaction with any information about the context
\item  \textbf{Context-only (Only Situation)}: 
Annotators only saw a description of the game and outcome for that round
\item \textbf{Context-based (Face+Situation)}: 
Annotators saw both the video and a description of the game and outcome
\end{itemize}

As BCI treats emotions judgments as a probability distribution over possible labels. This was estimated using the 20 annotations per video.  Following~\cite{anzellotti2021leveraging}, valence was discretized into five categories using the points of the Likert scale.

\subsection{Measuring Facial Expressivity}

Facial expressivity was quantified using visual features inspired by prior work~\cite{lei2023emotional}.
OpenFace 2.0~\cite{baltrusaitis2018openface} is used to extract 12 Facial Action Units (AUs)\footnote{The selected AUs are AU 1, 2, 4, 6, 7, 10, 12, 14, 15, 17, 25, and 26.}, focusing on frequently co-occurring AUs~\cite{stratou2017refactoring}. 
Optical flow is calculated using ZFace~\cite{jeni2015dense}, which tracks movements of 512 dense facial landmarks between frames. Head pose direction vectors, gaze direction vectors, and gaze angles are also included. 
Expressivity was calculated by combining facial, gaze, head movement, and optical flow metrics. 
Each metric was standardized and equally weighted.

We validate the calculated expressivity with human annotations for subset of dataset (24 videos). Human annotations use a 7-point Likert scale. The results show a correlation of 0.61 (p $<$ 0.001), comparable to the association found in~\cite{lei2023emotional}. 
This suggests that the automatic expressivity score is a reasonably proxy for the expressivity perceived by a human observer.

\subsection{Impact of Expressivity on Human Judgments}

BCI argues that context-based judgments reflect an equal integration of face-only and situation-only judgments of emotion.\footnote{BCI weighs the contribution of face or situation based on level of certainty in each channel; if annotators all agree the situation evokes sadness but are uncertain if the face is happy or sad, the situation receives more weight. All things being equal, the channels receive equal weight.} In contrast, Expression-Salience Hypothesis claims that highly expressive faces will capture attention, thereby assigning larger weight to the face.  

To test this, we analyze if the context-based judgments are closer to situation-only judgments or closer to face-only judgments, measured as a function of facial expressivity. 
Videos were grouped into tertiles based on their level of expressivity.
We then measure the proportion of context-based judgments that were closer to the face-only judgments versus situation-only judgments. 

\begin{figure}[h]
    \centering
    \includegraphics[width=1\linewidth]{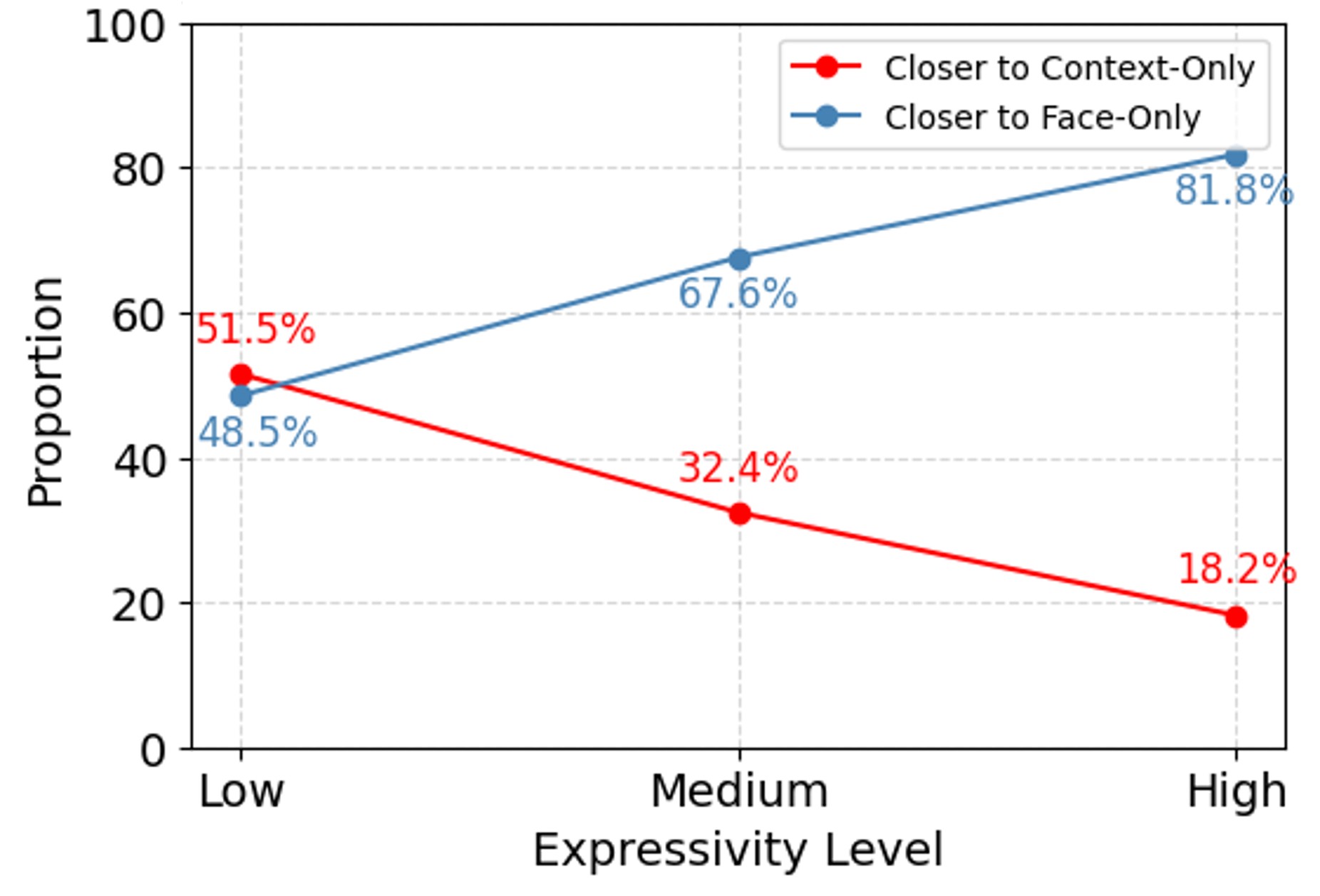}
    \caption{Proportion of context-based annotations that are best approximated by context-free versus context-only annotations (by facial expressivity).}
    \label{fig:proportion}
\end{figure} 

Figure~\ref{fig:proportion} shows clear support for for the Expression-Salience Hypothesis. When faces are highly expressive, most context-based judgments (82\%), are closer to perceptions of emotion from the face alone.  When faces are least expressive, context-based perceptions are equally influenced by the face and the situation. We next examine if this finding can be used to improve automatic expression recognition. 

\section{EXPERIMENT 1: Bayesian Cue Integration}
\label{sec:experiment1}

Although BCI was developed to explain human judgments, Han and colleagues showed it can be adapted for automatic context-based recognition~\cite{han2024knowledge}. Rather than using human annotations, they used standard facial expression recognition to predict context-free judgments and large langue models to predict context-only judgments. When integrated with BCI, the system quite accurately predicted human context-based judgments. Here, we see if these results can be improved further by incorporating expression salience. We test this for both basic emotion and valence recognition.

\subsection{Tasks and Evaluation Metrics}
We test our approach on both valence and basic emotion recognition. 
For valence, we evaluate the results using Mean Squared Error (MSE), Root Mean Squared Error (RMSE), and Pearson Correlation. Basic Emotion Recognition, treated as a distribution task, is evaluated using KL Divergence (KLD), RMSE, and Pearson Correlation. Specifically, we use KLD to measure the distance between the predicted and ground truth probability distributions~\cite{zhao2017approximating}.

\subsection{Bayesian Cue Integration with Salience Adjustment}
Section~\ref{sec:salience} showed that facial expressivity modulates how humans integrate these cues. To account for this, we introduce a salience adjustment mechanism into BCI.
The original BCI equation is shown in Eq.~\ref{eq:bayesian}~\cite{ong2015affective}, while the salience-adjusted version is defined in Eq.~\ref{eq:s-bayesian}, where \( w \) represents the weight assigned to the facial cue based on expressivity. \( P(e|f) \) represents the probability of emotion given only the face (Context-Free), \( P(e|c) \) represents the probability of emotion given only the game outcome (Context-Only), and \( P(e|c,f) \) represents Context-Based prediction.

\begin{equation}
P(e|c,f) \propto \frac{P(e|f)P(e|c)}{P(e)}
\label{eq:bayesian}
\end{equation}

\begin{equation}
P(e|c,f) \propto \frac{P(e \mid f)^w \cdot P(e \mid c)^{1-w}}{P(e)}
\label{eq:s-bayesian}
\end{equation}

We calculate the weight \( w \) based on the Expressivity Score (Section~\ref{sec:salience}) and rescale it to the range [0.5, 1.0] using a linear mapping, approximating the findings in Figure~\ref{fig:proportion}. 
Linear weighting is a common approach in cue integration models~\cite{ernst2002humans}, and in our case, it provides a practical adjustment method, aligning with prior psychological findings on attentional biases toward expressive faces~\cite{calvo2008detection, goel2024face}.

\subsection{Validation using Human Annotations}

We first evaluate salience adjustment by seeing if it improves BCI's fit to the human annotations: i.e., using the human annotations to estimate  \( P(e|f) \), \( P(e|c) \)  and \( P(e|c,f) \),  does Equation (2) better predict the distribution of \( P(e|c,f) \) than Equation (1). Table~\ref{table:bci_results} verifies that salience adjustment improves the accuracy of BCI for both valence and basic emotion prediction, averaging across the 100 videos.

\begin{table}[h!]
\centering

\scriptsize 
\resizebox{0.45\textwidth}{!}{%
\begin{tabular}{l|ccc}
\hline
\multicolumn{4}{c}{\textbf{Valence}} \\ \hline
                  & MSE(↓) & RMSE(↓) & Correlation(↑) \\ \hline
BCI (w/o Salience) & 0.199 & 0.446 & 0.743 \\  
BCI (w/ Salience)  & \textbf{0.108} & \textbf{0.328} & \textbf{0.870} \\ \hline
\multicolumn{4}{c}{\textbf{Basic Emotion}} \\ \hline
                  & KLD(↓) & RMSE(↓) & Correlation(↑) \\ \hline
BCI (w/o Salience) & 0.308 & 0.122 & 0.873 \\  
BCI (w/ Salience)  & \textbf{0.146}& \textbf{0.093} & \textbf{0.889} \\ \hline
\end{tabular}
}
\caption{Performance of BCI with Human-Generated Emotion}
\label{table:bci_results}
\end{table}

\subsection{Validation of Fully Automatic Recognition}
We next validate salience adjustment using automatic methods to estimate the emotion distributions and compare the automatic context-based prediction against the emotion distribution of the human context-based labels. 

We use pre-trained models to estimate \( P(e|f) \) (i.e., the context-free emotion predictions).  EmoNet~\cite{toisoul2021estimation} and Blueskeye (blueskeye.com) to estimate emotional valence.   Facet~\cite{littlewort2011computer} and EAC~\cite{zhang2022learn} estimate basic emotions. To compare with~\cite{han2024knowledge}, we also fine-tune a LSTM modeling following their protocol. We use GPT-4 to estimate \( P(e|c) \) (context-only emotion predictions) following~\cite{han2024knowledge}, specifically \textit{``gpt-4-o-mini"}. We prompt 20 times per description to ensure reliability and average the results.

\begin{table}[h]
\centering
\resizebox{0.45\textwidth}{!}{%
\begin{tabular}{@{}lccccc@{}}
\toprule
Model                     & Salience & MSE(↓)  & RMSE(↓) & Correlation(↑) \\ \midrule
EmoNet                   & -        & 0.516 & 0.718 & 0.425       \\
EmoNet+GPT &  w/o       & 0.481 & 0.694 & 0.528       \\
EmoNet+GPT  &  w/        & 0.474 & 0.688 & 0.584       \\
Blueskeye                & -        & 0.308 & 0.555 & 0.556       \\
Blueskeye+GPT &  w/o     & 0.274 & 0.523 & \textbf{0.653}       \\
Blueskeye+GPT &  w/     & \textbf{0.267} & \textbf{0.517} & 0.620       \\ \bottomrule
\end{tabular}
}
\caption{Performance of BCI with machine-generated valence}
\label{tab:valence_results}
\end{table}

\begin{table}[h]
\centering
\renewcommand{\arraystretch}{0.90} 
\resizebox{0.45\textwidth}{!}{%
\begin{tabular}{@{}lcccc@{}}
\toprule
Model          & Salience & KLD(↓)   & RMSE(↓)  & Correlation(↑) \\ \midrule
Facet          & -        & 2.437 & 0.203 & 0.059       \\
Facet+GPT  & w/o        & 2.065 & 0.199 & 0.238       \\
Facet+GPT  & w/        & 1.975 & 0.197 & 0.640       \\
EAC            & -        & 1.525 & 0.270 & 0.141       \\
EAC+GPT    & w/o       & 0.967 & 0.232 & 0.141       \\
EAC+GPT     & w/        & 0.938 & 0.190 & 0.603       \\
LSTM           & -        & 0.581 & 0.154 & 0.675       \\
LSTM+GPT    & w/o       & 0.537 & 0.139 & 0.725       \\
LSTM+GPT   & w/        &\textbf{0.347} & \textbf{0.109} & \textbf{0.725}       \\ \bottomrule
\end{tabular}
}
\caption{Performance of BCI with machine-generated basic emotions}
\label{tab:emotion_results}
\end{table}

The results show that salience adjustment consistently improves performance in both human and automatic emotion recognition tasks. 
By dynamically weighting facial and contextual cues, the proposed method better reflects human perception patterns and enhances performance

\section{EXPERIMENT 2: Vision-Language Models}
\label{sec:experiment2}
\begin{table*}[h]
\centering
\renewcommand{\arraystretch}{0.70} 
\resizebox{0.80\textwidth}{!}{%
\begin{tabular}{l|ccc|ccc}
\toprule
\textbf{Model} & \multicolumn{3}{c|}{\textbf{Valence}} & \multicolumn{3}{c}{\textbf{Basic Emotion}} \\ \cmidrule(lr){2-4} \cmidrule(lr){5-7}
 & \textbf{MSE(↓)} & \textbf{RMSE(↓)} & \textbf{Correlation(↑)} & \textbf{KLD(↓)} & \textbf{RMSE(↓)} & \textbf{Correlation(↑)} \\ \midrule
GPT-Free & 0.737 & 0.858 & 0.201 & 0.676 & 0.160 & 0.511 \\
GPT-Based (w/o Salience) & 0.600 & 0.775 & 0.584 & 0.549 & 0.128 & 0.600 \\
GPT-Based (w/ Salience) & \textbf{0.504} & \textbf{0.710} & \textbf{0.680} & \textbf{0.524} & \textbf{0.119} & \textbf{0.661} \\ 
\bottomrule
\end{tabular}
}
\caption{VLM-Context-Integration Results for Valence and Basic Emotion Tasks}
\label{tab:gpt-integration}
\end{table*}

With recent advancements in Vision-Language Models (VLMs)~\cite{lian2024gpt}, it is possible to fuse context and face information together in the same model, rather that using specialized models for each inference. As a last evaluation, we examine if VLMs could produce more accurate context-based emotion predictions by incorporating expression salience. 
We test this with \textit{gpt-4o-mini-2024-0718} due to its efficient reasoning capabilities and support for multimodal inputs.

\subsection{Setup and Frame Selection}
As GPT-4 cannot process video directly, we extract frames from videos and use them as input. While prior work~\cite{lian2024gpt} experimented with only 2–3 frames, we expand this range to 2–6 frames to evaluate the model’s performance across varying frame counts. We tested multiple frame counts within this range and found that performance was highest when using 4 frames.
After determining the number of frames, we uniformly sample from the original video, ensuring consistent representation of temporal information.

\begin{figure}[t]
    \centering
    \includegraphics[width=1\linewidth]{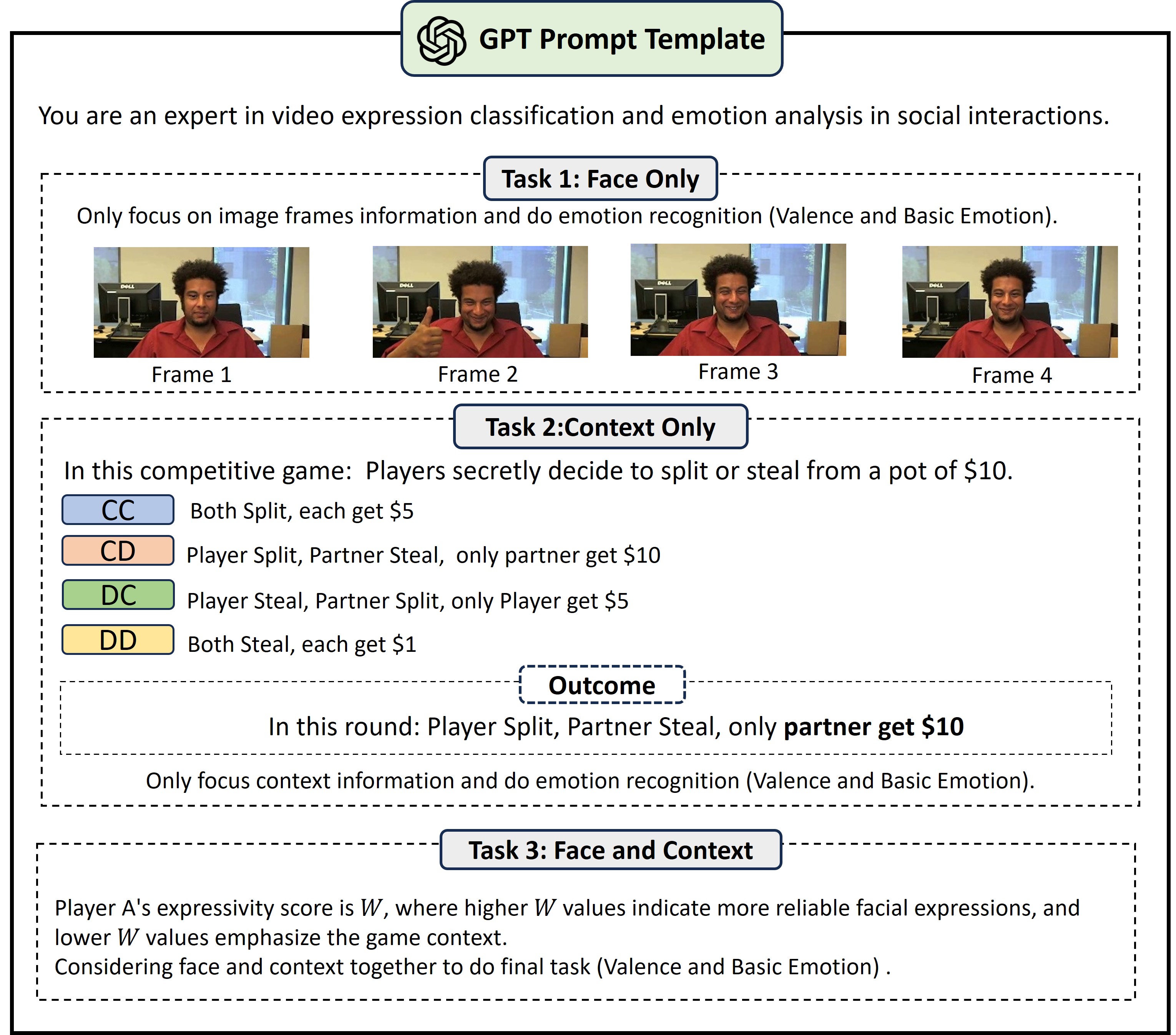}
    \caption{Salience Adjustment Prompt Template for GPT Visual Reasoning.}
    \label{fig:GPT-Based-Prompt}
\end{figure}

\subsection{Prompt Design}

To combine facial and contextual cues, we design a multi-step prompt process (Figure~\ref{fig:GPT-Based-Prompt}), inspired Chain-of-Thought reasoning~\cite{wei2022chain} in Vision-Language Models~\cite{zhang2024improve, ge2023chain}.
GPT performs three tasks sequentially:
1) Face-Only  emotion recognition (valence and basic emotion) based on the selected video frames; 
2) Context-Only emotion recognition using the joint outcome (e.g., CC or DC); and 
3) Face and Context integration based on both facial and contextual cues. The model also considers the facial expressivity score (\( W \)), where higher \( W \) values prioritize facial expressions, and lower \( W \) values emphasize game context.  


\subsection{Results: VLM-Context-Integration}

The result is shown in Table~\ref{tab:gpt-integration}. 
First, when comparing the Context-Free and Context-Based approaches for both tasks, we observe that the addition of contextual information consistently improves performance (consist with BCI approach). 
Furthermore, the adoption of the Salience Adjustment method further enhances performance, demonstrating the effectiveness of dynamically weighting facial and contextual cues to align with human perception patterns.

\section{Conclusion}

Our results provide strong support for the Expression-Salience Hypothesis and its utility for improving automatic emotion recognition. Within the prisoner's dilemma task, when the face shows strong emotions, observers attend more to the face and discount the results of the dilemma. Incorporating this ``salience adjustment" into BCI-based  recognition methods improved recognition accuracy. These improvements were found for both basic emotion recognition and valence recognition. Salience adjustment also improved the accuracy of an approach using a Visual Language Model. 

There are several next steps for this research. These results must be replicated on other social tasks. For example, the Split-Steal corpus contains a large number of smiles. Although this aligns with previous findings that smiles are the most common nonverbal signals in social settings~\cite{kraut1979social}, our results might not generalize to corpora with more diverse emotion distributions.
This study also focused exclusively on perceived emotion so a promising next step is to consider the accuracy of these perceptions compared to self-report as some research suggests context-based perceptions accurately reflect self-report~\cite{coan2007specific}. For example, automated methods have shown some success in distinguishing enjoyment smiles from other types of smiles~\cite{ambadar2009all}, and it would be useful to explore how automatically-recognized distinctions shape observer perceptions \textit{in context}.
Finally, our approach is a heuristic modification of BCI. The ``proper'' Bayesian solution would be to incorporate expressivity as an explicit variable and estimate its conditional effect on inferences via Bayes rule (which would require a larger corpus than we evaluated here). Again, extending these results to other domains will be a crucial next step. 

{\small
\bibliographystyle{ieee}
\bibliography{sample_FG2025}
}

\end{document}